\documentclass[conference]{IEEEtran}
\IEEEoverridecommandlockouts
\usepackage{cite}
\usepackage{amsmath,amssymb,amsfonts}
\usepackage{algorithmic}
\usepackage{graphicx}
\usepackage{textcomp}
\usepackage{xcolor}
\usepackage[hidelinks]{hyperref}
\usepackage{adjustbox}
\usepackage{amssymb}
\usepackage{booktabs}
\usepackage{multirow}
\usepackage{subcaption}
\usepackage{graphicx}
\usepackage[utf8]{inputenc}
\usepackage[T1]{fontenc}
\def\BibTeX{{\rm B\kern-.05em{\sc i\kern-.025em b}\kern-.08em
    T\kern-.1667em\lower.7ex\hbox{E}\kern-.125emX}}

\def\our{Emotional Normalizing Flow}

\begin{document}

\title{Modeling Uncertainty in Personalized Emotion Prediction with Normalizing Flows}

\author{\IEEEauthorblockN{
Piotr Miłkowski$^{\dagger 1}$, Konrad Karanowski$^{\dagger 1}$, Patryk Wielopolski$^1$, Jan Kocoń$^1$,  Przemysław Kazienko$^1$, 
Maciej Zięba$^{1, 2}$}
\IEEEauthorblockA{\textit{$^1$ Department of Artificial Intelligence, Wrocław University of Science and Technology, Poland}}
\IEEEauthorblockA{\textit{$^2$ Tooploox, ul. Tęczowa 7, 53-601 Wrocław, Poland}}
\IEEEauthorblockA{\footnotesize{\texttt{\{piotr.milkowski,konrad.karanowski,patryk.wielopolski,jan.kocon,kazienko,maciej.zieba\}@pwr.edu.pl}}}} 

\maketitle

\begin{abstract}
Designing predictive models for subjective problems in natural language processing (NLP) remains challenging. This is mainly due to its non-deterministic nature and different perceptions of the content by different humans. It may be solved by Personalized Natural Language Processing (PNLP), where the model exploits additional information about the reader to make more accurate predictions. However, current approaches require complete information about the recipients to be straight embedded. Besides, the recent methods focus on deterministic inference or simple frequency-based estimations of the probabilities. In this work, we overcome this limitation by proposing a novel approach to capture the uncertainty of the forecast using conditional Normalizing Flows. This allows us to model complex multimodal distributions and to compare various models using negative log-likelihood (NLL). In addition, the new solution allows for various interpretations of possible reader perception thanks to the available sampling function. We validated our method on three challenging, subjective NLP tasks, including emotion recognition and hate speech. The comparative analysis of generalized and personalized approaches revealed that our personalized solutions significantly outperform the baseline and provide more precise uncertainty estimates. The impact on the text interpretability and uncertainty studies are presented as well. The information brought by the developed methods makes it possible to build hybrid models whose effectiveness surpasses classic solutions. In addition, an analysis and visualization of the probabilities of the given decisions for texts with high entropy of annotations and annotators with mixed views were carried out.
\end{abstract}


\section{Introduction}

\def\thefootnote{$\dagger$}\footnotetext{These authors contributed equally to this work.}

Human affective states, including emotions, strongly depend on the individual, the stimulant eliciting them, and the associated context \cite{barrett2017emotions}. Therefore, the reasoning of a person's perception based on machine learning bears a significant degree of uncertainty. It refers to the reaction to any content, including text reading. We can say that disagreements in human textual inferences are inherent \cite{10.1162/tacl_a_00293}. Most solutions to subjective problems in natural language processing (NLP), like recognition of emotions, hate speech, sarcasm, sense of humor, sentiment, and many others, rely on generalized perspectives. They consider only text and its single generalized interpretation. Then, the commonly used solution is to simplify multiple distinct views, i.e., annotations provided by many annotators using majority voting or other methods to achieve a sole perception. Overall, we can identify two sources of uncertainty: (1) humans, who are unsure and imprecise in their annotations (this is a hidden factor), and (2) a community of annotators. The latter refers to discrepancies between people in understanding the problem, and perception of a given text \cite{beck2020representation, davani2022dealing, troiano2021emotion}. The standard measures for inter-rater agreement are Krippendorff's alpha \cite{krippendorff2011} or  Fleiss' kappa \cite{fleiss1971measuring}.
However, they provide only a single value characterizing the set of all annotations for all texts. Yet another (3) source of uncertainty: the trained model itself. It means that the model is not capable of precisely learning about concepts (what is \textit{joy} or \textit{hate speech}?) and relations from the available learning samples. This leads to errors and proximate reasoning.
Simultaneously, emotions can be considered multidimensional objects, which requires multi-task learning \cite{Milkoski2022Multitask} and further complicates the problem of uncertainty modeling.
Most of the proposed approaches for subjective modeling in the NLP domain focus on deterministic predictions. In this work, we propose to enrich the family of emotional methods by introducing \our{} -- an entirely probabilistic framework that utilizes conditional Normalizing Flows to model uncertainty. We postulate to represent the considered tasks as multivariate regression problems and represent the distribution of the outputs with conditional flows. This approach allows us to model complex multimodal distributions of multidimensional outputs. The experiments and validation were carried out on emotion detection (ten tasks) and hate speech (two tasks). We examine various choices of flow models and compare their performance with the mixture of Gaussians, showing the superiority of \our{} compared to the selected baseline. Moreover, we show that incorporating personalization into our model leads to better distribution adjustment measured with negative log-likelihood (NLL) value.

To summarize, the contributions of this work are as follows:
\begin{itemize}
\item We introduce a novel approach for probabilistic modeling in subjective NLP-based problems; 
\item We examine the impact of personalization on the quality of the model and show that in most of the considered experimental cases, additional information about the reader leads to better probability adjustment; 
\item We show that our approach outperforms the standard baseline that utilizes a mixture of Gaussians;  
\item We propose a hybrid approach utilizing Normalizing Flows and personalization that outperform previous models.
\end{itemize}

\section{Related work}

Initial work on emotion recognition in the text was based mainly on frequency analysis of words defined in lexicons of emotions \cite{strapparava2008learning,tabak2016comparison}. These lexicons contained words with assigned categories of basic emotions, e.g., joy, anger, sadness \cite{plutchik1980general,ekman1992argument}. Emotions occurring most frequently at the lexical level were then assigned to the entire text. With the development of text classification methods based on machine learning, datasets containing texts manually annotated with emotions began to emerge \cite{oberlander2018analysis,kocon2018context,kocon2019propagation,kocon2019recognition,kocon2019multilingual,demszky2020goemotions,kocon2021aspectemo,srivastava2023beyond}. Due to annotators' subjective perception of emotions, and thus low inter-annotator agreement, it was common to assign emotion labels to text based on majority voting \cite{oberlander2018analysis,demszky2020goemotions}. Based on such prepared data, text classification models were trained. Initially, such models as SVM \cite{mohammad2012emotional}, BiLSTM, and GRU \cite{abdul2017emonet} were used. Currently, transformer-based models such as BERT perform best in the task of emotion recognition \cite{hsu2018socialnlp, demszky2020goemotions}. The aforementioned approaches require data for which the inter-annotator agreement is high. However, there are some data sets such as Wiki-Detox \cite{wulczyn2017ex}, Sentimenti \cite{kocon2019recognition} or Measuring Hate Speech \cite{kennedy2020constructing}, which contain an annotator identifier linked to their affective annotation. They also include multiple annotations for a given text from multiple annotators. For such data, new personalized approaches have recently been developed, in which the context of the annotator is taken into account in the model learning process \cite{akhtar2020modeling,Kocon2021ipm,mireshghallah2021useridentifier,kanclerz2021controversy,milkowski2021acl,ngo2022studemo,kanclerz2022if,bielaniewicz2022deep,ferdinan2023personalized,mieleszczenko2023capturing,kocon2023differential}. This makes it possible, for example, to answer the question of what emotions a particular text evokes for a particular user. 
Recent method proposals also focus on neuro-symbolic approaches to explain decisions made \cite{cambria2022senticnet}, usage of large-scale pre-trained language model (PLM) for prompt-based classification tasks such as sentiment analysis and emotion detection \cite{mao2022biases}, using recently popular large language models (LLMs) \cite{amin2023wide}, or methods of complex persona attribute extraction \cite{zhu2023paed}.
However, the methods mentioned above do not model the uncertainty associated with the community's subjective perception of emotions and the degree of indecision of the annotators themselves.

In this paper, to model uncertainty described in the Introduction
we adapt the concept of Normalizing Flows. The best-known Normalizing Flow models such as NICE \cite{dinh2014nice}, RealNVP \cite{dinh2016density}, MAF \cite{papamakarios2018masked}, and CNF \cite{grathwohl2018ffjord} were originally used for density estimation and image generation tasks. These models were further extended and used as components for more sophisticated tasks or even for other domains of applications. In Computer Vision, there were proposed models such as RegFlow \cite{zikeba2020regflow} for probabilistic future location prediction, Flow Plugin Network \cite{wielopolski2021flow}, PluGeN \cite{WolczykPMZWKS22}, and StyleFlow \cite{AbdalZMW21} models for conditional image generation. For the tabular data, recently, TreeFlow \cite{wielopolski2022treeflow} was proposed that utilizes a combination of tree-based models with conditional Normalizing Flows to estimate uncertainty for uni- and multi-variate regression problems. 
In terms of Natural Language Processing and Normalizing Flows, only Discrete Flow \cite{TranVADP19} was proposed to model character-level datasets using Normalizing Flows dedicated to the discrete data. 
To the best of our knowledge, no probabilistic approach has been proposed to model distributions of uncertainty in personalized natural language processing, and our \our{} is the first probabilistic model proposed for multi-task prediction of personalized emotions.

\section{Background}
\label{sec:background}

\paragraph{Generalized and Personalized Approach to Subjective NLP Problems.}

In the classic approach to the task of text classification or regression, we assume a training set of the form $\mathcal{D}=\{(\mathbf{t}_i, \mathbf{y}_i)\}_{i=1}^N$, where $\mathbf{t}_i \in \mathcal{T}$ is the $i$-th text document and $\mathbf{y}_i$ is its annotation. 
However, many NLP tasks, such as recognizing emotions in a text or detecting hate speech, can be subjective because each person perceives these phenomena. This leads to a situation when we can have more than one annotation $\mathbf{y}$ for the same text $\mathbf{t}$, as different people may annotate the same texts differently. Therefore, a training set is in the form of $\mathcal{D}=\{(\mathbf{t}_i,\mathbf{p}_i, \mathbf{y}_i)\}_{i=1}^N$, where $\mathbf{y}_i$ is the annotation given by person $\mathbf{p}_i \in \mathcal{P}$ for text $\mathbf{t}_i \in \mathcal{T}$. 

\begin{figure}[!ht]
\centering
\begin{subfigure}[b]{0.5\textwidth}
   \includegraphics[trim={0 2cm 0 2cm}, width=1\linewidth]{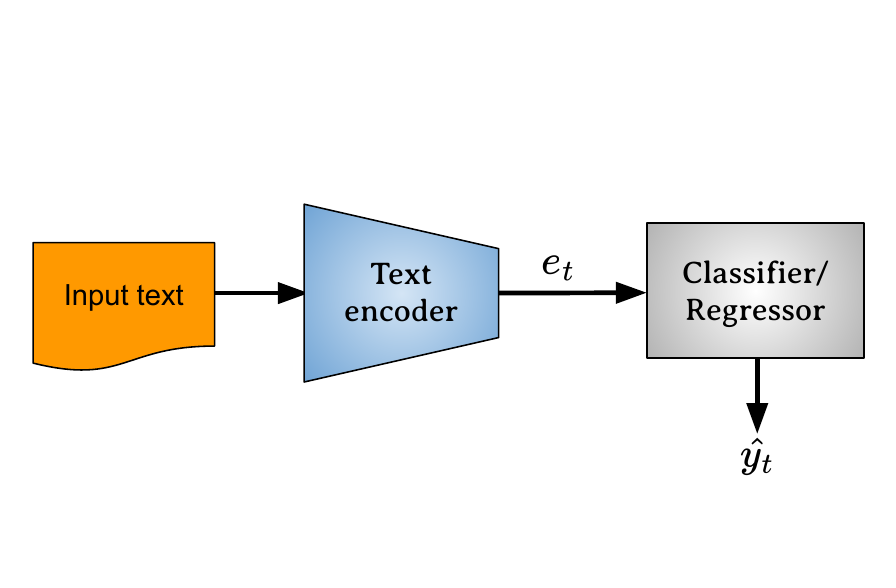}
   \caption{Generalized deterministic model.}
\end{subfigure}

\begin{subfigure}[b]{0.5\textwidth}
   \includegraphics[trim={0 1cm 0 1cm}, width=1\linewidth]{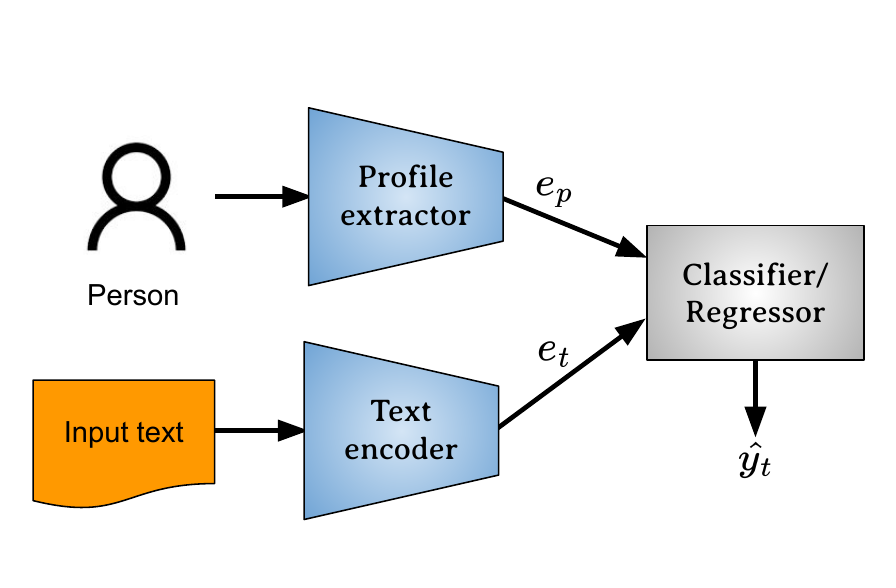}
   \caption{Personalized deterministic model.}
\end{subfigure}

\caption[Comparison of deterministic models]{
    Comparison of (a) generalized and (b) personalized deterministic models. (a) consists of two parts: a text encoder (language model) that creates text embedding $e_t$ and a classifier or regressor (mostly fully-connected layer) that provides prediction $\hat{y}$. This approach is not considered suitable for subjective NLP tasks, like emotion recognition, because it does not respect the individual perception of the text. Model (b) fixes this problem by adding a profile extractor in the form of user representation $e_p$. It allows human individual characteristics to be included in the inference process.
    Both models are deterministic, giving us limited, spolight information about subjective tasks.
}
\label{fig:deterministic-models}
\end{figure}

One approach to subjective tasks in NLP is the so-called \textit{generalized}. It assumes that the model predicts the result based solely on the text and returns the same prediction for every user. Generalized models usually consist of two parts: \textbf{text encoder} (language model), which creates \textbf{text representation} $e_t$ and \textbf{classifier} or \textbf{regressor} (usually fully-connected layer) that gives prediction $\hat{y}$.
However, recent studies \cite{mireshghallah2021useridentifier, kocon2021learning, KAZIENKO2023} show that this approach should not be considered correct, as adding information about the annotator significantly improves model quality and yields better results. The approach that combines information about the text and the human is so-called \textit{personalized}. Compared to the generalized, personalized model adds another component called \textbf{profile extractor}, that creates \textbf{human representation} $e_p$. The comparison of generalized and personalized approaches is shown in Fig~\ref{fig:deterministic-models}.

There are few existing architectures \cite{mireshghallah2021useridentifier, kocon2021learning} utilizing this fact. Still, all of them are deterministic, meaning none model uncertainty as a direct optimization of negative log-likelihood.

\paragraph{Normalizing Flows.}

Normalizing Flows \cite{rezende2015variational} are a class of generative models that enables estimation of the uncertainty of prediction thanks to the access to log probability function and thus enable direct optimization of negative log-likelihood (NLL). The goal of the model is to transform base distribution $p_{U}(\mathbf{u})$ (usually Gaussians with independent components) to the complex distribution of the data $p_{Y}(\mathbf{y})$ using a series of $K$ invertible functions that can be written as $\mathbf{u}=\mathbf{f}_K \circ \dots \circ \mathbf{f}_1(\mathbf{y})$. For that purpose, Normalizing Flows utilize the change-of-variable formula and then the NLL $\mathbf{y}$ is given by
\begin{equation}
\log p_{Y}(\mathbf{y}) = \log p_{U}(\mathbf{u}) - \sum_{k=1}^K \log  \left| \det \frac{\partial \mathbf{f}_n}{\partial \mathbf{z}_{k-1}} \right|.
\end{equation}
To specify the exact Normalizing Flow model, we need to define transformations $\mathbf{f}_1, \dots, \mathbf{f}_K$. Here, multiple models were proposed such as NICE \cite{dinh2014nice}, RealNVP \cite{dinh2016density}, MAF \cite{papamakarios2018masked} or Continuous Normalizing Flows \cite{grathwohl2018ffjord}.

\begin{figure}[!ht]
\centering
\begin{subfigure}[b]{0.5\textwidth}
   \includegraphics[width=1\linewidth]{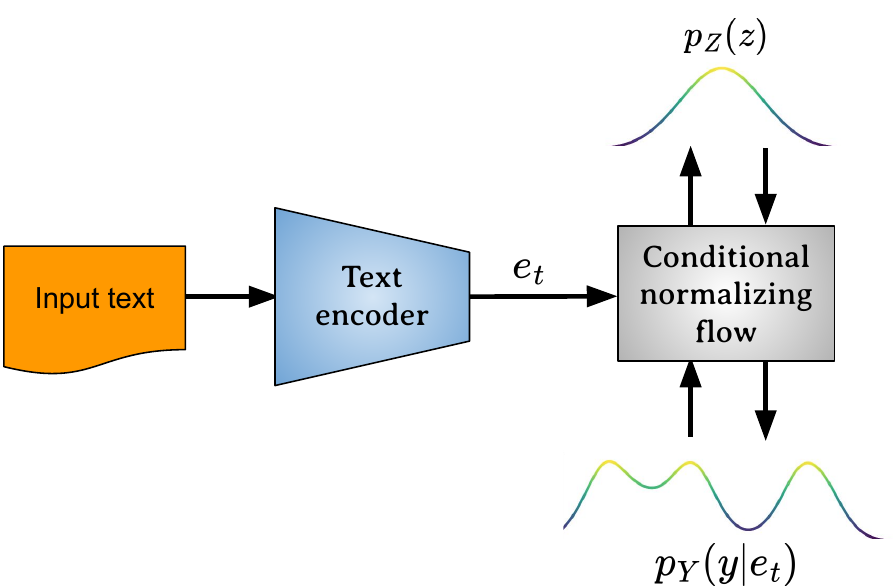}
   \caption{Generalized flow-based probabilistic model.}
\end{subfigure}

\begin{subfigure}[b]{0.5\textwidth}
   \includegraphics[width=1\linewidth]{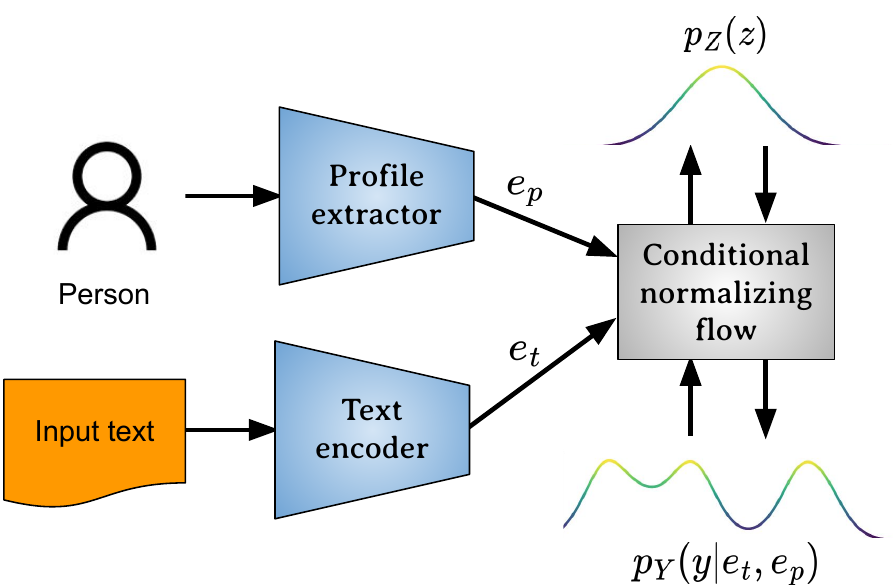}
   \caption{Personalized flow-based probabilistic model.}
\end{subfigure}

\caption[Comparison of probabilistic models]{
Comparison of (a) generalized and (b) personalized flow-based probabilistic models. 
    Model (a), as in the case of generalized deterministic, uses only information about the text. However, unlike it, it models the conditional probability distribution $p_Y(\mathbf{y}|\mathbf{e}_t)$ using Normalizing Flow.
    Model (b) extends the concept of the personalized deterministic model in a similar way to (a) but it exploits representations of both the text $e_t$ and user $e_p$ to model conditional probability distribution $p_Y(\mathbf{y}|\mathbf{e}_p, \mathbf{e}_t)$ for the subjective output predictions representing emotions. 
}
\label{fig:probabilistic-models}
\end{figure}

\section{Our approach}

In this section, we introduce \our{} - the probabilistic model for subjective uncertainty modeling in the NLP domain. The general schema of the proposed approach is provided in Fig.~\ref{fig:probabilistic-models}. The model is composed of \emph{Profile extractor} that is responsible for creating the representation of the person, $\mathbf{e}_p$, and \emph{Text encoder} that creates embedding $\mathbf{e}_t$ directly from the input text. Both components can be represented by various models (trainable and fixed), and we elaborate on this further in this section. 

The extracted vectors $\mathbf{e}_p$ and $\mathbf{e}_t$ are further delivered to the conditional flow represented by the complex transformation function $\mathbf{f}(\cdot)$. The role of the function is to transform multivariate regression outputs $\mathbf{y}$ to  $\mathbf{z}$ that represents the variable in the base space, assuming given vectors, $\mathbf{e}_p$ and $\mathbf{e}_t$. Formally, we have $\mathbf{z}=\mathbf{f}(\mathbf{y},\mathbf{e}_p, \mathbf{e}_t)$, where $\mathbf{f}$ is  invertible with respect to $\mathbf{\mathbf{y}}$, $\mathbf{y}=\mathbf{f}^{-1}(\mathbf{z}, \mathbf{e}_p, \mathbf{e}_t)$. Moreover, the complex transformation $\mathbf{f}$ can be decomposed into a sequence of simple functions, $\mathbf{z}=\mathbf{f}_K \circ \dots \circ \mathbf{f}_1(\mathbf{y}, \mathbf{e}_p, \mathbf{e}_t)$, where the $K$ is number discrete transformations. With such assumptions, the probability distribution for $\mathbf{y}$ that represents the distribution over the regression outputs can be calculated using the formula:
 
\begin{equation}
p_{Y}(\mathbf{y}|\mathbf{e}_p, \mathbf{e}_t)=  p_{Z}(\mathbf{z}) \cdot \prod_{k=1}^K \left| \det \frac{\partial \mathbf{f}_k}{\partial \mathbf{z}_{k-1}} \right|, 
\end{equation}
where $\mathbf{z}_0, \dots, \mathbf{z}_K$ are intermediate steps after discrete transformations, assuming $\mathbf{z}_0:=\mathbf{y}$, and $\mathbf{z}_K:=\mathbf{z}$. $p_{Z}(\mathbf{z})$ is the assumed base distribution for $\mathbf{z}$ with the known density function, usually represented by Gaussian. 
Consequently, we have direct access to the density function for that conditional distribution. Therefore we can calculate the likelihood function for a set of input-output pairs to evaluate the quality of the model. We can sample an infinite number of output values assuming given inputs and interpret the results.  

The proposed model can quickly adapt to the problems without personalization, simply skipping $\mathbf{e}_p$ conditioning in the flow. Our approach is independent of the conditional Normalizing Flow type, and we experimentally compare the performances of the most popular models. We follow the methodology of incorporating conditional components described in \cite{wielopolski2021flow}. 

\paragraph{Profile extractor.}

Vector $\mathbf{e}_p$ contains information about the user specific to the personalization architecture used. This can include information such as the deviation of responses from the majority voice, metadata about the user, user identifier \cite{mireshghallah2021useridentifier}, the correlation of the text's context with historical evaluations, or other features unique to the recipient of the text. It also can be randomly initialized and tuned during the learning process by backpropagation \cite{kocon2021learning}.

\paragraph{Text encoder.}

In the case of $\mathbf{e}_t$ vector, text representation is implemented using Transfomer language models. An attentional weight is assigned for a given text input, divided into individual tokens. The assigned values are then used to calculate the weighted sum of the resulting vectors \cite{vig2019analyzing}. It is possible to fine-tune the language model using the loss function of the final model.

\paragraph{Training the model.} To trained \our{} we use the dataset $\mathcal{D}=\{(t_n,p_n, \mathbf{y}_n)\}_{n=1}^N$, composed of $t_n$ textual input, $p_n$ features of the person, and corresponding subjective annotation $y_n$ given by the person $p_n$ for text $t_n$. We train our model directly by optimizing the negative log-likelihood function:

\begin{equation}
\mathcal{L}= - \sum_{n=1}^N  \log p_{Y}(\mathbf{y}_n|\mathbf{e}_{n,p}, \mathbf{e}_{n, t}),
\end{equation}
where $\mathbf{e}_{n,p}$ is a vector, that represents profile of the person $p_n$, and $\mathbf{e}_{n,t}$ is an embedding of the text $t_n$. The model can be trained in a two-stage mode or end-to-end paradigm depending on the form of \emph{Profile extractor} and \emph{Text encoder}. In the first case, the embeddings  $\mathbf{e}_{p}$ and $\mathbf{e}_{t}$ are extracted in the first stage, and parameters of the flow are trained while optimizing $\mathcal{L}$. Alternatively, suppose the \emph{Profile extractor} or \emph{Text encoder} are represented by differentiable architectures. In that case, the entire system can be optimized end-to-end, directly minimizing the negative log-likelihood function.

\section{Experiments}

In this subsection, we evaluate our approach on a set of challenging datasets, investigating the impact of adding contextual information about the person in the model. Moreover, we compare flow-based probabilistic models to a simple Gaussian Mixture Model. Then, we compare our solution to the deterministic models using sampling from flow and discretization. Finally, we mix deterministic and probabilistic approaches to create a hybrid model. 

\subsection{Datasets}

\paragraph{Wikipedia-Detox.}

The Wikipedia Detox project has created a crowd-sourced dataset that contains one million annotations covering 100,000 discussions of page edits on Wikipedia \cite{wulczyn2017ex}. These were often filled with toxic statements, verbal aggression, and even personal attacks. Each comment was annotated by about ten annotators provided by the Crowdflower service.

The collection containing toxic statements consists of 160,000 texts. It includes a binary determination of toxicity (where: 0 = non-toxic, 1 = toxic), as well as a rating from -2 to 2 (where: 2 = very healthy, 0 = neutral, and -2 = very toxic).

Sets for personal attacks and verbal aggression consist of 100,000 of the same comments. In addition to the binary marks for aggression (0 = neutral or friendly comment, 1 = aggressive or attacking), aggression is put on a scale analogous to toxicity from -2 to 2 (where: 2 = very friendly, 0 = neutral, and -2 = very aggressive). Personal attacks are divided into types: quoting, recipient, third party, or another type of attack. In addition to texts shared between these collections, the same applies to annotators. Thus, we can use knowledge from one collection to benefit from it in another or a collective approach. Those willing to participate in the study also completed questionnaires so that we have demographic information about them available.

\paragraph{Emotion Simple.}

This collection consists of 100 texts marked on 10 scales by 5,365 annotators \cite{milkowski2021personal}. Texts are opinions posted on websites. This gives 53.65 annotations per text and 1.69 markings from a single user. The texts were rated for eight basic emotions (sadness, anticipation, joy, fear, surprise, disgust, trust, and anger) and emotional arousal on a scale from 0 to 4 for each dimension. In addition, the tenth aspect rated is the valence expressed on a scale from -3 to 3 (where -3 = negative, 0 = neutral, and 3 = positive). In the set of individuals with two marks, those with three or more annotations also appear.

\paragraph{Emotion Meanings.}

In \cite{wierzba2021emotion}, a huge collection containing 6,000 assessed word collocations was prepared and published. It contains dimensions and scales analogous to the Emotion Simple collection -- the basic emotions from Robert Plutchik's Wheel of Emotions \cite{plutchik1991emotions}.

The scale of the collection makes it one of the most interesting and, simultaneously, the most difficult for personalizing emotion detection. It has 303,143 annotations from 16,101 people who participated in the study. Each collocation has been evaluated 50.67 times, and a single annotator has an average of 18.83 annotations.

The difficulty in working with these data is also because these are not full-fledged textual statements containing context but just two words. An example item from the collection: "colorful beads". Annotator data include information such as gender, age, education, size of residence, relationship status, income, or political views.

\subsection{Setups}

The dataset was divided into training, validation, and testing splits. Users and texts were not mixed between sets to bring the evaluation as close to the real-world scenario. Each experiment consisted of 10-fold cross-validation, and obtained results were averaged. 
Statistical significance tests were performed: t-test with Bonferroni correction to address the problem of multiple comparisons. In the tables within the rows, comparisons were made between models without and with personalization. Bold indicates the best result, and underline indicates the absence of statistically significant differences for each dataset. Within the ``Type`` column, the best probabilistic model type or no significant difference between the two was similarly marked for each dataset separately.

\paragraph{Baselines.}
We have three reference points. To check the impact of personalization, we compared personalized models with a baseline that uses only textual information (TXT-Baseline); it is a generalized approach. To investigate the impact of normalizing flows, we compared them with a Gaussian Mixture Model to have a reference point in the form of another, less complex probabilistic method. Finally, we compared our method with deterministic approaches. 

\paragraph{Models for conditional normalizing flows.}

In our experimental evaluation, we consider \our{} with various types of conditional normalizing flows. For single-dimensional datasets: \textit{Wikipedia Detox: Toxicity}, \textit{Wikipedia Detox: Aggression} and \textit{Wikipedia Detox: Attack}, we used MAF (\textbf{maf}) and CNF (\textbf{cnf}). For multi-dimensional \textit{Emotions Meanings} and \textit{Emotions Simple}, we used two extra flows: RealNVP (\textbf{real\_nvp}) and NICE (\textbf{nice}). We compared the results against the baseline that uses mixtures of Gaussians to model the probability (\textbf{gmm}).  

\paragraph{Models for personalization.}
We investigate three approaches to respect the personalization context: OneHot, HuBi-Formula, and HuBi-Medium \cite{kocon2021learning}. They are confronted with TXT-Baseline (generalized, non-personalized) that does not contain any information about the annotator. We exploit LaBSE \cite{feng-etal-2022-language} as a language model in every experiment.

\subsection{Experimental scenarios}

\paragraph{Experiment 1 - Comparison of generalized and personalized solutions in the probabilistic approach.}

The first approach verifies the performance of \our{} with fixed hyperparameters on multiple data sets and tasks: \textit{Wikipedia Detox: Toxicity}, \textit{Wikipedia Detox: Aggression}, \textit{Wikipedia Detox: Attack}, \textit{Emotion Meanings} and \textit{Emotion Simple}. We also verified the ability of the proposed \our{} to transfer knowledge between thematically similar multidimensional text labels. For this purpose, the \textit{Wikipedia Detox: Aggression} and \textit{Wikipedia Detox: Attack} datasets were joined, as they contain annotations for the same texts performed by the same annotators. As a result, we obtained a dataset with multi-dimensional labels. This experiment aimed to examine the effect of personalization models on the prediction of probability distributions, thus verifying whether the additional information provided to the model reduces its uncertainty and comparing Normalizing Flows to Gaussian Mixture Model. 

\paragraph{Experiment 2 - Investigating the effect of hyperparameters selected for personalization and \our{} methods on the most difficult dataset.}
The second approach was to verify the maximum possible reduction of model uncertainty by tuning the model hyperparameters to a given set and checking which normalizing-flow model obtained the best results. Due to limited resources, we decided to perform this experiment using only \textit{Emotion Meanings} dataset. The parameters that we tuned were: the number of hidden features, number of layers, number of blocks per layer, dropout probability, batch normalization within layers, batch normalization between layers, learning rate, the size of hidden layers used to prepare user embeddings, and the size of the output of these embeddings.

\paragraph{Experiment 3 - Comparison of probabilistic and deterministic approaches.}
To compare with classical methods \cite{Milkoski2022Multitask}, which are deterministic, it was necessary to prepare conversions of the \our{} output to the form of exact values. Included in the body of the paper is the application of two best normalizing flows (RealNVP and CNF) for multidimensional datasets (\textit{Aggression \& Attack} [classification task] and \textit{Emotion Simple} [regression task]).

For the first type of task, each text or text-user pair was sampled using an iterative method. In the preparation step, we increased the number of samples in the test part of the dataset so that the value from 0 to 1 with a step of 0.1 for the class was tested as a possible context. Iterations were done twice for values of 0 and 1 in the opposite class. Next, an exponential was applied to the 44 probabilities of the resulting sample (22 per class for each text). Within the values for the opposite sampling, (e.g., [0.5, 0] and [0.5, 1]) of a given dimension were summed, and then for each stopper (0.0, 0.1, ..., 1.0) divided by the sum of all values for the dimension. If the probability mass prevailed on the side from 0.0 to 0.5, it was considered that the class was not assigned and vice versa for the other part of the axis.

It was impossible for a 10-dimensional set for the regression task to sample each possible dimension in all values separately because of the number of possible combinations. Each item from the test subset was replicated 100 times containing random real values from 0 to 1 in each class. Majority voting was then conducted to determine the most likely response for the scale of each dimension. In the collection, each dimension had a value analogous to the slider setting during annotation. For this reason, the task was treated as an ordinary regression, and the resulting values were rounded to the nearest possible position. This assumption was used for both values from the deterministic and probabilistic approaches.

\paragraph{Experiment 4 - Hybrid approach (utilizing knowledge from the text and uncertainty modeling).}
The combined approach, hereafter referred to as hybrid, was done in two steps. In the first, the learned \our{} models were sampled in the same way as in Experiment 3, but for all the texts in the collection. Then, the network input was extended to the deterministic model with an additional feature. A vector containing the resulting probabilities for each text was entered along with its embeddings and, in the case of approaches with personalization, the user profile. This vector contained all the values from the sampling, and no additional mathematical operations were performed on it.

\subsection{Results}

\paragraph{Results of Experiment 1.}
The first experiment proved that adding personalization reduces the uncertainty of probabilistic models, Tab.~\ref{tab:experiment_1}. For \textit{Wikipedia Detox} datasets (\textit{Aggression}, \textit{Attack} and \textit{Toxicity}), all personalized models received significantly lower negative log-likelihood values compared to the non-personalized TXT-Baseline. For all three tasks, the best architecture was HuBi-Medium combined with CNF.
For \textit{Aggression \& Attack} dataset, personalization improved most cases' results. The best results were obtained by OneHot combined with RealNVP and HuBi-Formula combined with CNF. 
In the case of \textit{Emotion Simple} and \textit{Emotion Meaning}, personalization also reduced model uncertainty in most of the cases. For both datasets, the best results were obtained by the HuBi-Medium model combined with RealNVP.
It is worth noting that compared to Gaussian Mixture Model, Normalizing Flows always obtain lower negative log-likelihood values. It suggests that target variables, i.e., emotions, have complex distributions, and using a simple probabilistic approach is not enough.

\paragraph{Results of Experiment 2.}
In the second experiment, we carried out hyperparameter tuning on the most challenging dataset: \textit{Emotion Meanings}, Tab.~\ref{tab:experiment_2}. All possible combinations of hyperparameters were considered when performing the grid search. The results seem to confirm earlier speculations about MAF's better ability to deal with multidimensional problems compared to other approaches. Moreover, none of the variants indicated the benefit of using text alone as input. 

\paragraph{Results of Experiment 3.}
In the third experiment, we compared results obtained by deterministic models and \our{}, Tab.~\ref{tab:experiment_3}. It was carried out on two datasets: combined \textit{Wikipedia Detox:  Aggression \& Attack} and \textit{Emotions Simple}. We also decided to use only two Normalizing Flow Models that performed the best on both of these datasets: RealNVP and CNF.  

In the case of  \textit{Aggression \& Attack} dataset, the results obtained by deterministic models were better for every architecture, including the non-personalized one. 
In the case of \textit{Emotion Simple} dataset, the results obtained by probabilistic models significantly outperformed deterministic models. The best model was a combination of HuBi-Medium and CNF.
This result seems to confirm that the probabilistic approach is especially effective on complex multi-dimensional tasks. 

\paragraph{Results of Experiment 4.}
In the fourth experiment, we mixed the deterministic approach with the probabilistic, to create a hybrid model, Tab.~\ref{tab:experiment_4}.
In both \textit{Aggression \& Attack} and \textit{Emotion Simple} tasks, the results obtained by the hybrid approach outperformed previous methods by a large margin. 
For \textit{Aggression \& Attack} dataset, the best model turned out to be HuBi-Medium with CNF. 
For \textit{Emotion Simple} dataset, HuBi-Medium with both RealNVP and CNF performed comparably well.
The results of this experiment prove, that adding information about the model uncertainty makes big difference in the inference process, and helps to predict better for difficult and subjective tasks.

\begin{table}[t]
    \centering
    \begin{adjustbox}{width=\columnwidth}
    \begin{tabular}{llrrrr} 
\toprule
Dataset          & Type               & TXT-Baseline      & OneHot             & HuBi-Formula       & HuBi-Medium         \\ 
\midrule
Toxicity         & maf                & 0.0702          & -0.0197          & \textbf{-0.0947} & 0.0053            \\
                 & cnf                & 0.1231          & -0.0707          & -0.1202          & \textbf{-0.1378}  \\
                 & gmm                & 0.6422          & 0.6164           & 0.5829           & \textbf{0.5072}   \\
\midrule
Aggression       & maf                & \underline{0.1705}  & \underline{0.1695}   & \textbf{0.0526}  & 0.0859            \\
                 & cnf                & 0.1685          & 0.0978           & 0.0180           & \textbf{-0.0431}  \\
                 & gmm                & 0.8948          & \underline{0.7783}   & \underline{0.7841}   & \textbf{0.7446}   \\ 
\midrule
Attack           & maf                & 0.1669          & 0.1180           & \underline{-0.0229}  & \underline{-0.0250}   \\
                 & cnf                & 0.1474          & 0.0811           & -0.0427          & \textbf{-0.0950}  \\
                 & gmm                & 0.7512          & 0.7318           & 0.7105           & \textbf{0.6911}   \\ 
\midrule
Aggression \& Attack & maf          & \textbf{-1.3788} & N/A & -0.3966          & -0.3834    \\
                   & nice         & -0.8678          & \textbf{-1.1281}     & -1.0524          & -1.0914    \\
                   & real\_nvp    & -3.3482          & \textbf{-3.6181}     & -3.0235          & -1.7208    \\
                   & \textbf{cnf} & -3.5113          & -2.7339              & \textbf{-3.7002} & -2.1357    \\
                   & gmm          & 3.1729           & \textbf{2.4028}      & 2.5673           & 2.6858     \\
\midrule
Emotion Meanings & maf                & 0.5337          & \textbf{-0.0135} & 0.8525           & 0.1936            \\
                 & nice               & -1.9099         & -1.8707          & \textbf{-2.0283} & -1.4792           \\
                 & \textbf{real\_nvp} & \underline{-5.4775} & -2.9377          & \underline{-5.5189}  & \textbf{-5.6377}  \\
                 & cnf                & -3.7186         & -1.9640          & -3.4632          & \textbf{-4.8458}  \\
                 & gmm                & 5.9559          & 5.4858           & 4.8034           & \textbf{4.4719}   \\ 
\midrule
Emotion Simple   & maf                & \underline{1.9130}  & 2.6398           & 2.6393           & \underline{1.9314}    \\
                 & nice               & 2.4254          & \textbf{1.8163}           & 2.3502           & 1.9613            \\
                 & \textbf{real\_nvp} & 2.5583          & 1.7845           & 2.3726           & \textbf{1.4355}   \\
                 & cnf                & 2.1220          & \textbf{1.8197}           & 2.3347           & 1.9702            \\
                 & gmm                & 4.2706          & \textbf{3.7496}  & 4.2312           & 4.0910            \\ 

\bottomrule
    \end{tabular}
    \end{adjustbox}
    \caption{Experiment 1 - negative log-likelihood values for all datasets, without hyperparameter tuning.}
    \label{tab:experiment_1}
\end{table}

\begin{table}[h]
    \centering
    \begin{adjustbox}{width=\columnwidth}
    \begin{tabular}{llrrrr}
\toprule
Dataset          & Type         & TXT-Baseline & OneHot              & HuBi-Formula       & HuBi-Medium         \\ 
\midrule
Emotion Meanings & \textbf{maf} & -11.5380   & \textbf{-14.0785} & -10.6476         & -12.1934          \\
                 & nice         & 4.2167     & \textbf{-2.0243}  & -1.0978          & -1.9208           \\
                 & real\_nvp    & -2.3509    & -4.5813           & -5.2833          & \textbf{-7.0285}  \\
                 & cnf          & -1.9623    & -3.9381           & \textbf{-5.2247} & -4.9381           \\
                 & gmm          & 12.6564    & 9.6047            & \textbf{7.8908}  & 8.4545          \\
\bottomrule
    \end{tabular}
    \end{adjustbox}
    \caption{Experiment 2 - negative log-likelihood values for Emotion Meanings dataset, with hyperparameter tuning.}
    \label{tab:experiment_2}
\end{table}

\begin{table}[t]
    \centering
    \begin{adjustbox}{width=\columnwidth}
    \begin{tabular}{llcccc}
\toprule
Dataset                                                                     & Method                       & TXT-Baseline   & OneHot                 & HuBi-Formula           & HuBi-Medium              \\ 
\midrule
Aggression \& Attack & deterministic  & \underline{0.5874} & \underline{0.5954}         & 0.7354                 & \textbf{0.7847}          \\ 
{[}Macro F-1] & discrete(real\_nvp)          & 0.4829         & 0.4682                 & 0.5860                 & \textbf{0.6788}          \\
   & \textbf{discrete(cnf)}  & 0.5189   & 0.4738  & 0.6397 & \textbf{0.7374}  \\ 
\midrule
Emotion Simple & deterministic  & 0.3936 & 0.5403  & 0.5574 & \textbf{0.5822}  \\ 
{[}R2] & \textbf{discrete(real\_nvp)} & 0.4472 & \underline{0.6535} & \underline{0.6582} & \textbf{0.6835}  \\
        & discrete(cnf)    & 0.4428 & 0.6274 & 0.6685  & \textbf{0.7005}  \\ 
\bottomrule
    \end{tabular}
    \end{adjustbox}
    \caption{Experiment 3 - Comparison of classification and regression using the classical deterministic method and the result of sampling \our{}.  Macro F-1 score for \textit{Aggression \& Attack} and R-squared for \textit{Emotion Simple} datasets.}
    \label{tab:experiment_3}
\end{table}

\begin{table}[t]
    \centering
    \begin{adjustbox}{width=\columnwidth}
    \begin{tabular}{llcccc}
\toprule
Dataset  & Method & TXT-Baseline & OneHot & HuBi-Formula & HuBi-Medium \\ 
\midrule
Aggression \& Attack & deterministic  & \underline{0.5874} & \underline{0.5954} & 0.7354 & \textbf{0.7847}  \\ 
{[}Macro F-1] & hybrid(real\_nvp)  & 0.8479 & \underline{0.8254} & \underline{0.8233} & \textbf{0.8743} \\
              & \textbf{hybrid(cnf)} & 0.8693  & 0.9052  & 0.8400  & \textbf{0.9553} \\ 
\midrule
Emotion Simple & deterministic  & 0.3936 & 0.5403 & 0.5574 & \textbf{0.5822}  \\ 
{[}R2] & \underline{hybrid(real\_nvp)}  & 0.4722 & 0.7149 & 0.7237 & \textbf{0.7376}  \\
        & \underline{hybrid(cnf)} & 0.4867 & \underline{0.6899} & \underline{0.69223} & \textbf{0.7388}  \\
\bottomrule
    \end{tabular}
    \end{adjustbox}
    \caption{Experiment 4 - Comparison of the classical deterministic approach and hybrid models, which in addition use probabilistic knowledge. Macro F-1 score for \textit{Aggression \& Attack} and R-squared for \textit{Emotion Simple} datasets.}
    \label{tab:experiment_4}
\end{table}

\section{Conclusions}

In this paper, we proposed a novel \our{} approach to personalized NLP that opens up new perspectives on predicting reader behavior in a non-deterministic way. 
From the perspective of psychology and the variability of emotion sensation over time, the problem of emotion recognition is one of the most difficult and subjective tasks facing NLP. People do not perceive their emotions as zero-one, and most of the attempts so far classified their feelings in this way. The presented probabilistic approach based on normalizing flows provides more complex information about the uncertainty and diversity of possible emotional states. A comparative analysis of models for emotion recognition without and with personalization indicated that new methods are also effectively applicable in a non-deterministic setup. The generalized, non-personalized solution generates a completely different concentration of probability mass, directed toward a quantitative approach. Personalization can shift the view of the problem in a contextual way by dedicating reasoning to a single user.
Finally, we showed that adding information about model uncertainty significantly improves the ability to predict complex and subjective behaviors such as recognizing hate speech or emotions in a text. The hybrid model we created significantly outperformed the previous methods, becoming a new state-of-the-art on two very challenging tasks.
Our future work will focus on applications of our approach to some other tasks such as active or reinforcement learning.

\section*{Limitations}
One important issue related to the nature of normalizing flows is their ability to convert probabilities to disambiguate uncertain answers. At the moment, there are no reference datasets available that contain text and annotator information simultaneously with multiple markings of the same text by the same person. This is due to cost constraints in preparing such data. However, we have conducted experiments on datasets with different characteristics in which (1) one person marked several hundred texts [\textit{Wikipedia Detox Datasets}] and (2) one text was evaluated dozens of times by different people [\textit{Emotions Simple} and \textit{Emotion Meanings} datasets]. In order to address the problem mentioned in the introduction, one text should have $N$ annotations from the same person, e.g., a few days apart. If we gain access to or prepare such a dataset, we would be happy to conduct in-depth studies on it.

Due to the language of one of the datasets being different from English, a multilingual model was used to embed the text. This decision was made in order to allow for direct comparisons and cross-referencing. This would have added an unnecessary layer to the already relatively complex problem that was addressed. It is possible to experiment with other language models as well using the source codes provided\footnote{\tiny{\url{https://github.com/CLARIN-PL/personalized-emotion-prediction-with-normalizing-flows}}}.

\section*{Acknowledgments}
This work was financed by 
(1) the National Science Centre, Poland, project no. 2021/41/B/ST6/04471;  
(2) Contribution to the European Research Infrastructure 'CLARIN ERIC - European Research Infrastructure Consortium: Common Language Resources and Technology Infrastructure', 2022-23 (CLARIN Q);
(3) the Polish Ministry of Education and Science, CLARIN-PL; 
(4) the European Regional Development Fund, as a part of the 2014-2020 Smart Growth Operational Programme, projects no. POIR.04.02.00-00C002/19, POIR.01.01.01-00-0923/20, POIR.01.01.01-00-0615/21, and POIR.01.01.01-00-0288/22; 
(5) the statutory funds of the Department of Artificial Intelligence, Wroclaw University of Science and Technology;
(6) the Polish Ministry of Education and Science within the programme “International Projects Co-Funded”;
(7) the European Union under the Horizon Europe, grant no. 101086321 (OMINO). However, the views and opinions expressed are those of the author(s) only and do not necessarily reflect those of the European Union or the European Research Executive Agency. Neither the European Union nor European Research Executive Agency can be held responsible for them.
The work conducted by Maciej Zieba and Patryk Wielopolski was supported by the National Centre of Science (Poland) Grant No. 2021/43/B/ST6/02853.

\bibliography{anthology, custom}
\bibliographystyle{ieeetr}

\appendix

\section{Appendix}
\label{sec:appendix}

\subsection{Personalization architectures}

\begin{figure}[h]
    \centering
    \includegraphics[width=\linewidth]{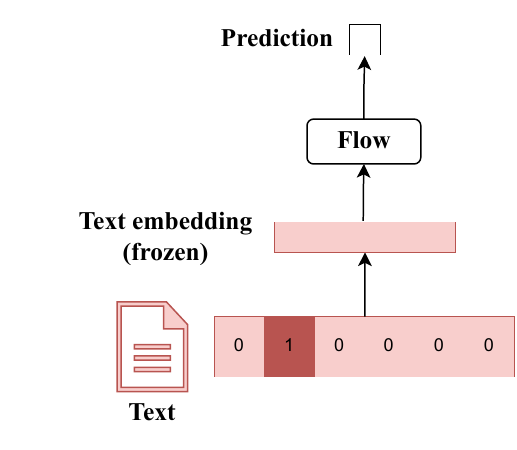}
    \caption{TXT-Baseline architecture utilizing normalizing flows.}
    \label{fig:txt_baseline}
\end{figure}

\begin{figure}[h]
    \centering
    \includegraphics[width=\linewidth]{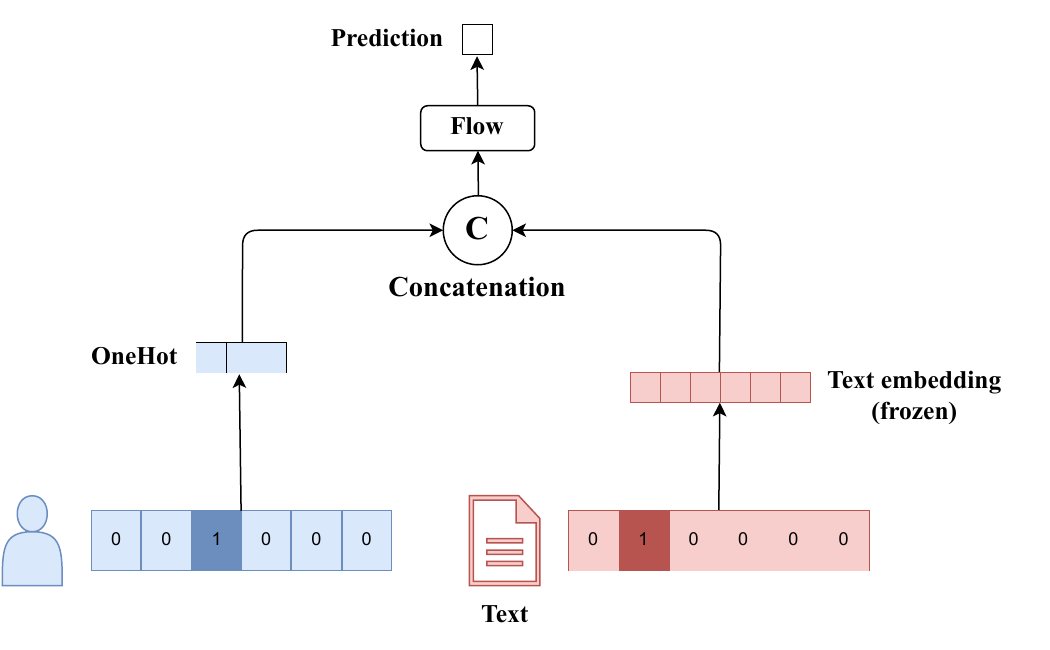}
    \caption{OneHot architecture utilizing normalizing flows.}
    \label{fig:onehot}
\end{figure}

\begin{figure}[h]
    \centering
    \includegraphics[width=\linewidth]{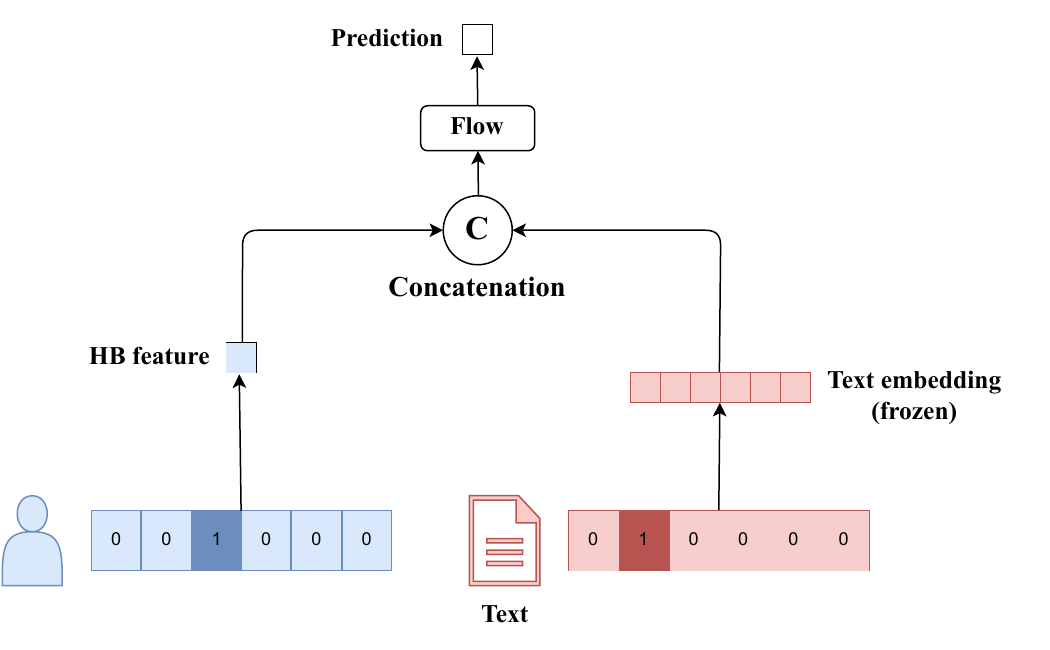}
    \caption{HuBi-Formula architecture utilizing normalizing flows.}
    \label{fig:hubi_formula}
\end{figure}

\begin{figure}[h]
    \centering
    \includegraphics[width=\linewidth]{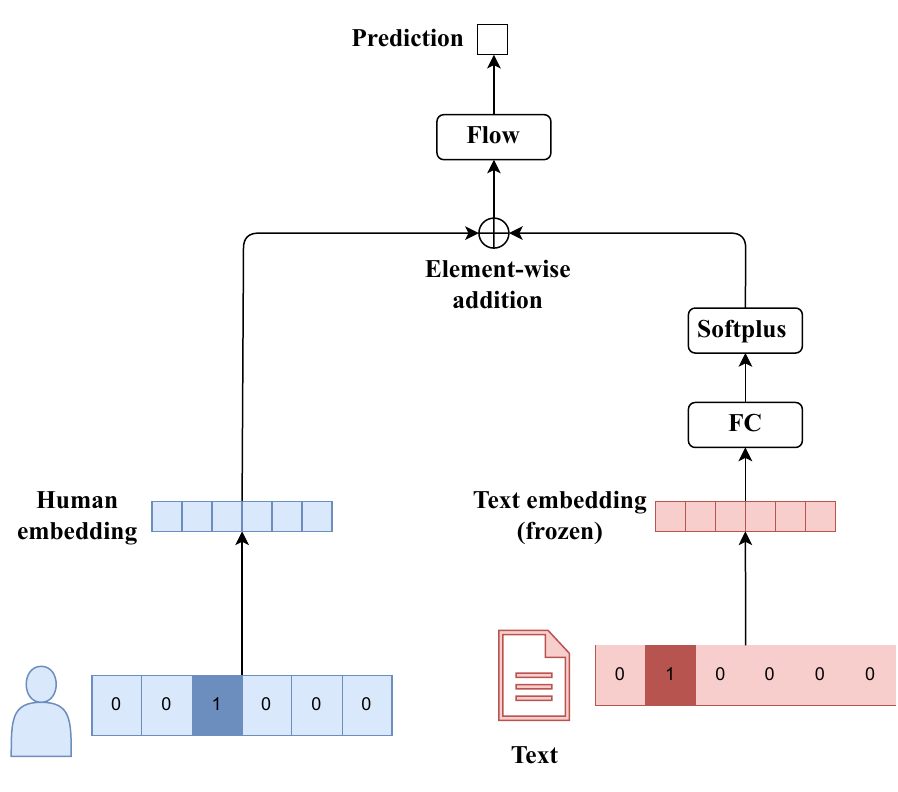}
    \caption{HuBi-Medium architecture utilizing normalizing flows.}
    \label{fig:hubi_medium}
\end{figure}

Architectures for personalization have been modified to return a probability distribution instead of the probability of a class. Input features are passed to the normalizing flow as context. We are dealing with four architectures:
\begin{itemize}
    \item Baseline (Fig.~\ref{fig:txt_baseline}) - the input is just an embedding of text
    \item OneHot (Fig.~\ref{fig:onehot}) - the user represented as a one-hot vector is concatenated to the text embedding
    \item HuBi-Formula (Fig.~\ref{fig:hubi_formula}) - the deviation of the user's response is taken as its representative feature
    \item HuBi-Medium (Fig.~\ref{fig:hubi_medium}) - annotation-based learned user embedding combined with text embeddings provides the context
\end{itemize}

\subsection{Implementation details}

Experiments were performed using the code provided in the “Anonymous”.

Grid search for hyperparameters of Normalizing Flows in experiment 2 was performed with the values specified in Tab.~\ref{tab:app_hyperparameters_flows} and Tab.~\ref{tab:app_hyperparameters_models}.

\begin{table}[h]
\begin{tabular}{l|ll}
Hyperparameter            & \multicolumn{2}{l}{Values}                   \\
\hline
hidden features           & \multicolumn{2}{l}{{[}2, 4, 6, 8{]}}         \\
\hline
num layers                & \multicolumn{2}{l}{{[}1, 2, 3, 4, 5{]}}      \\
\hline
num. blocks per layer     & \multicolumn{2}{l}{{[}1, 2, 3, 4{]}}         \\
\hline
dropout probability       & \multicolumn{2}{l}{{[}0.0, 0.1, 0.2, 0.4{]}} \\
\hline
batch norm within layers  & \multicolumn{2}{l}{{[}True, False{]}}        \\
\hline
batch norm between layers & \multicolumn{2}{l}{{[}True, False{]}}  
\end{tabular}
\caption{Hyperparameters for Normalizing Flows and their possible values during experiment 2. Note that for MAF, the \textit{dropout probability} hyperparameter was not used at all.}
\label{tab:app_hyperparameters_flows}
\end{table}

\begin{table}[h]
\centering
\begin{adjustbox}{width=\columnwidth}
\begin{tabular}{l|l|l}
Hyperparameter & Values for TXT-Baseline \& OneHot \& HuBi-Formula & Values for HuBi-Medium \\ 
\hline
embedding dim. & 50                                                & 50                     \\ 
\hline
hidden dim.    & 50                                           &      [128, 256, 512, 786]                  \\ 
\hline
output dim.    &       -         &        [128, 256, 512, 768]                \\ 
\hline
learning rate  & {[}1e-5, 1e-4{]}                                  & {[}1e-5, 1e-4{]}      
\end{tabular}
\end{adjustbox}
\caption{Hyperparameters for training and architectures, and their possible values during experiment 2.}
\label{tab:app_hyperparameters_models}
\end{table}

For all experiment purposes, we used a machine with AMD Ryzen Threadripper PRO 3955WX 16-Core Processor CPU, 2 x NVIDIA GeForce RTX 3090 GPUs, and 256 GB RAM.

\subsection{Visualization of probabilities}

\begin{figure}
    \centering
    \begin{subfigure}[b]{0.24\textwidth}
        \centering
        \includegraphics[width=\textwidth]{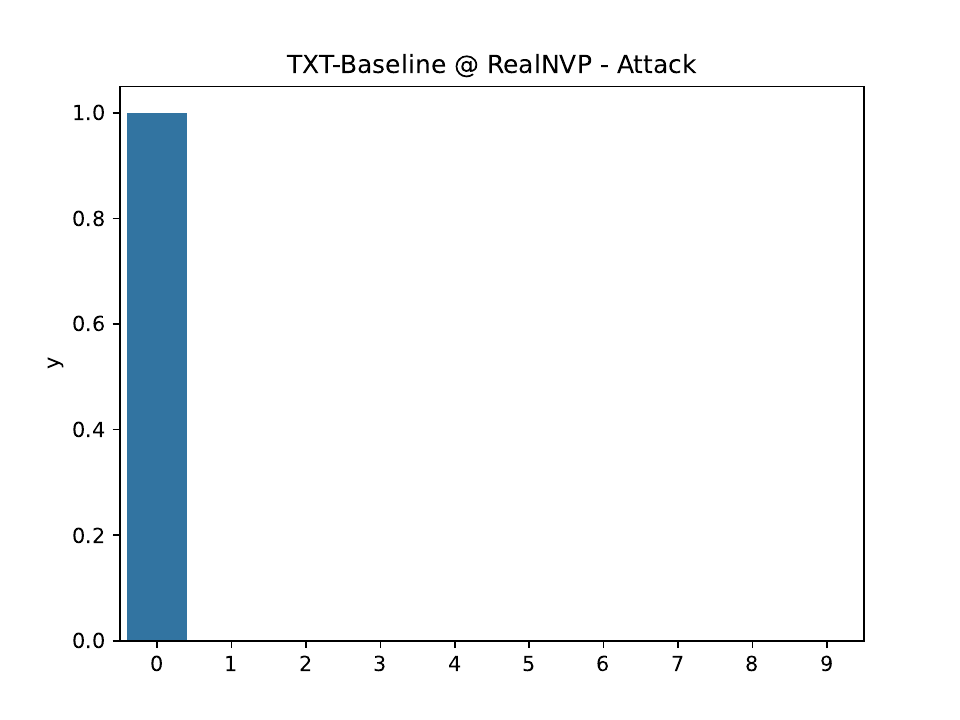}          
        \caption[]%
        {{\small}}    
    \end{subfigure}
    \hfill
    \begin{subfigure}[b]{0.24\textwidth}  
        \centering 
        \includegraphics[width=\textwidth]{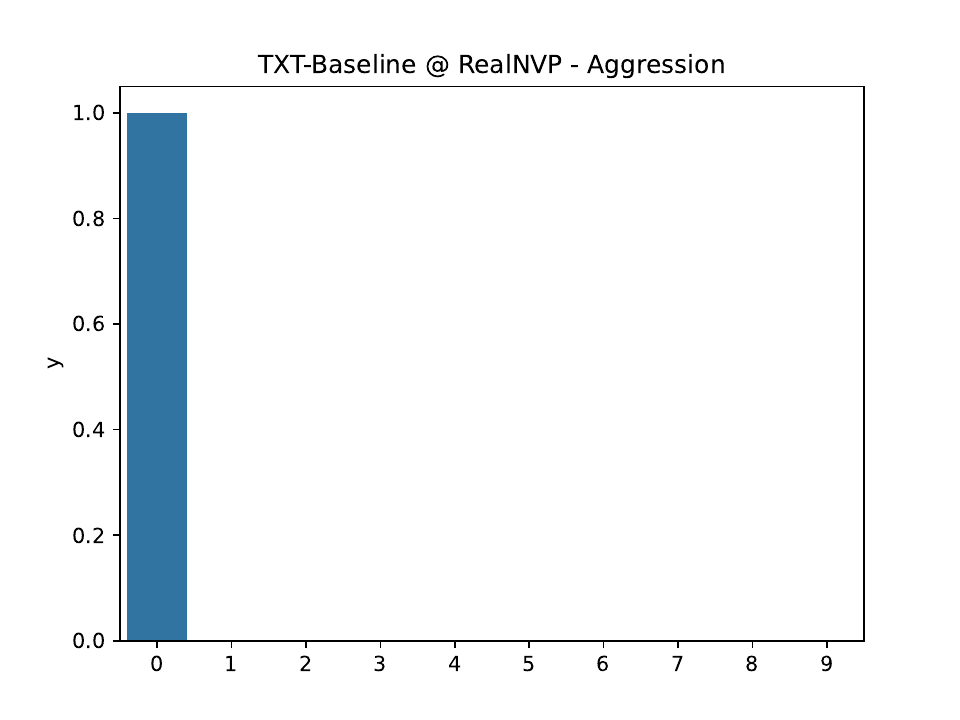}          
        \caption[]%
        {{\small}}    
    \end{subfigure}
    \vskip\baselineskip
    \begin{subfigure}[b]{0.24\textwidth}   
        \centering 
        \includegraphics[width=\textwidth]{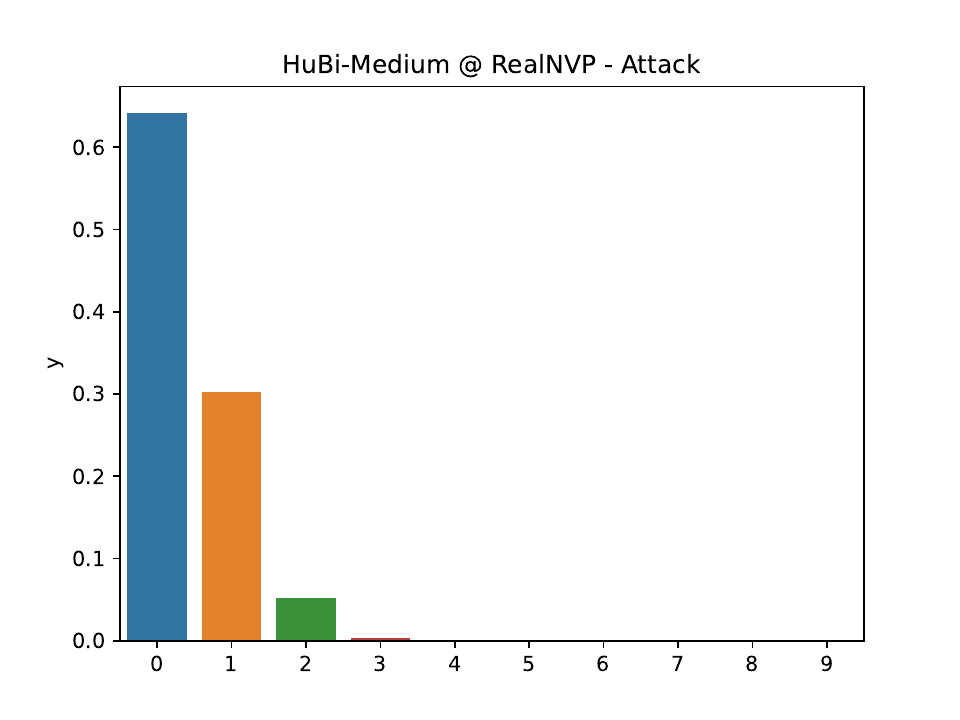}
      \caption[]%
        {{\small}}    
    \end{subfigure}
    \hfill
    \begin{subfigure}[b]{0.24\textwidth}   
        \centering 
        \includegraphics[width=\textwidth]{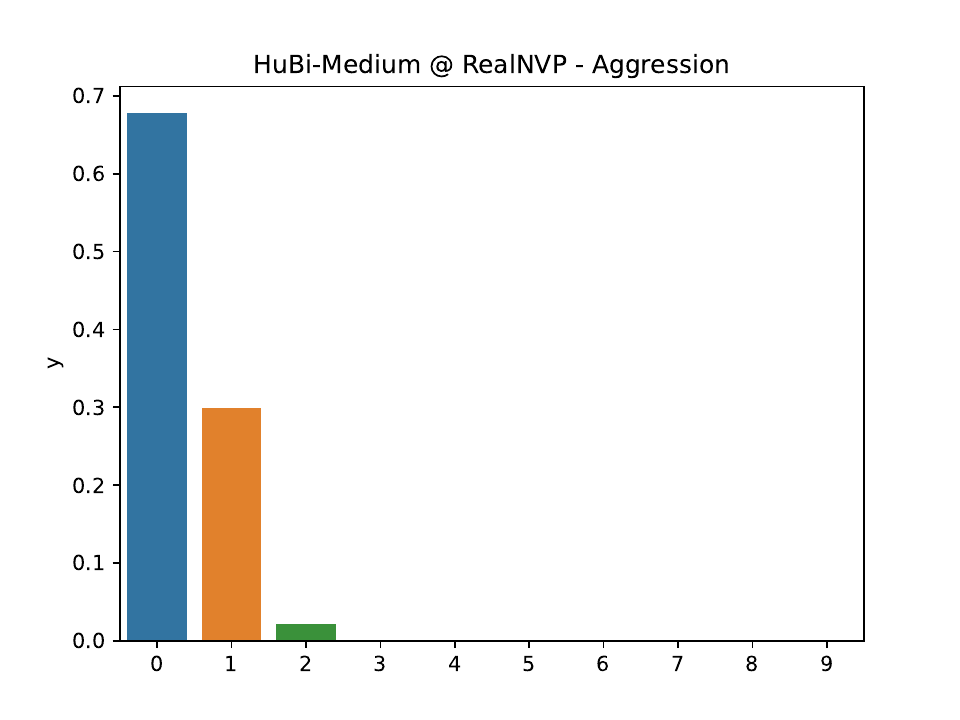}
        \caption[]%
        {{\small}}    
    \end{subfigure}
    \caption[]
    {Visualizations of the waveform of the probability function for RealNVP with different architectures and for Attack and Aggression.} 
    \label{fig:realnvp_histo}
\end{figure}

\begin{figure}
    \centering
    \begin{subfigure}[b]{0.24\textwidth}
        \centering
        \includegraphics[width=\textwidth]{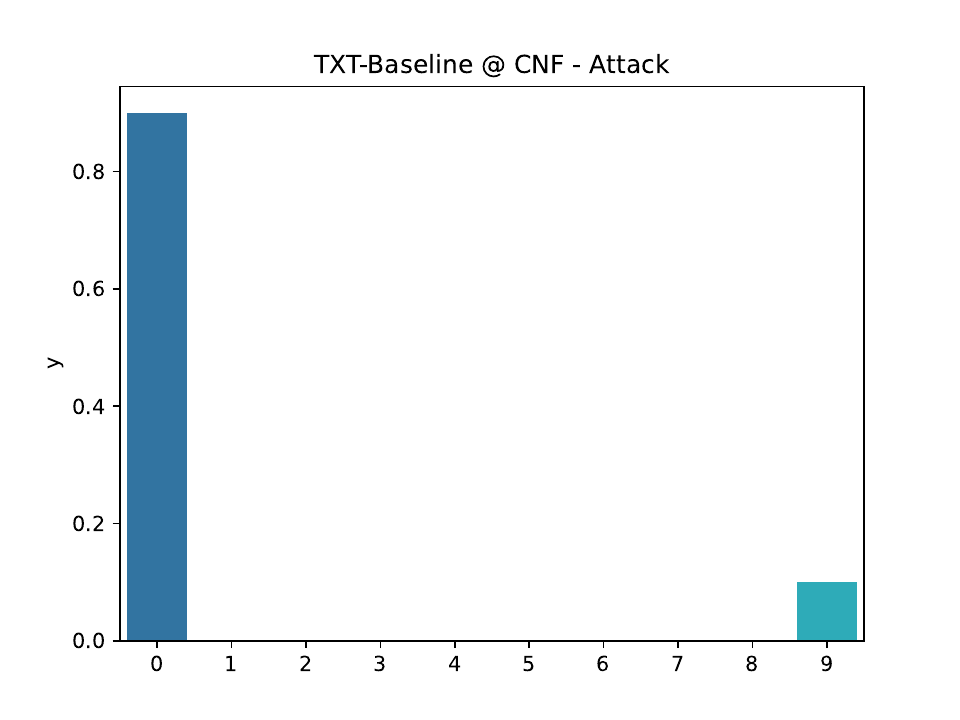}          \caption[Network2]%
        {{\small}}    
    \end{subfigure}
    \hfill
    \begin{subfigure}[b]{0.24\textwidth}  
        \centering 
        \includegraphics[width=\textwidth]{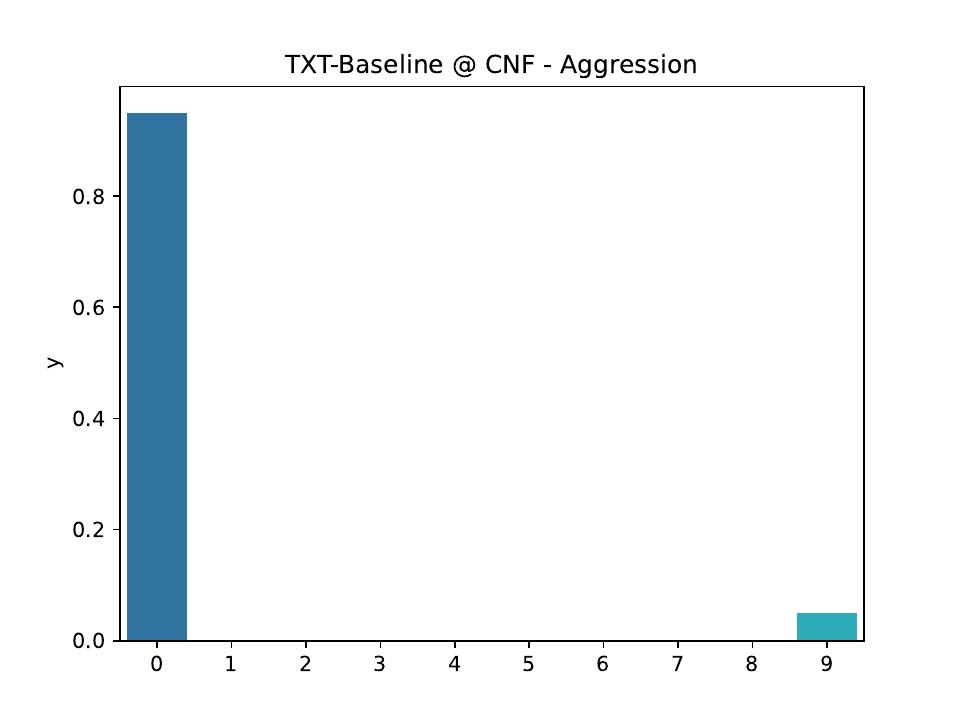}          \caption[]%
        {{\small}}    
    \end{subfigure}
    \vskip\baselineskip
    \begin{subfigure}[b]{0.24\textwidth}   
        \centering 
        \includegraphics[width=\textwidth]{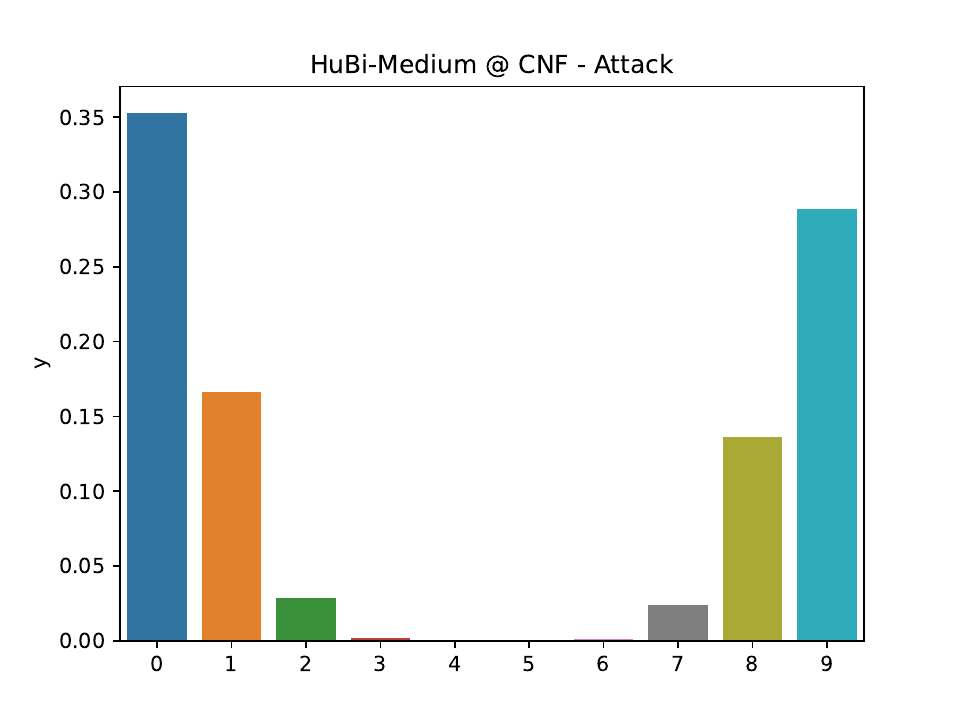}
      \caption[]%
        {{\small}}    
    \end{subfigure}
    \hfill
    \begin{subfigure}[b]{0.24\textwidth}   
        \centering 
        \includegraphics[width=\textwidth]{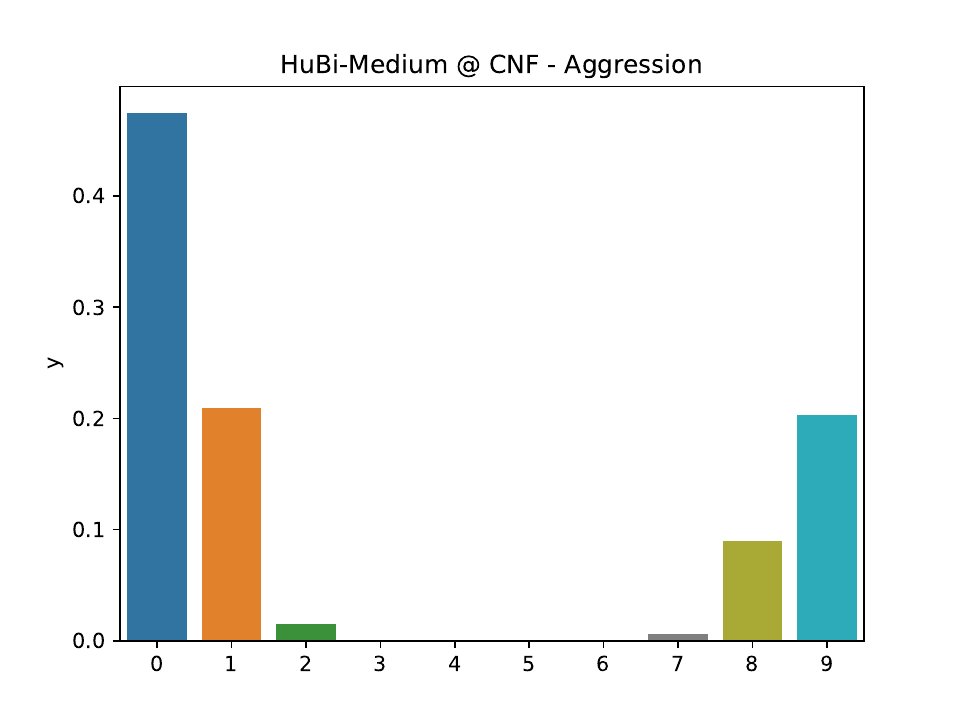}
        \caption[]%
        {{\small}}    
    \end{subfigure}
    \caption[]
    {Visualizations of the waveform of the probability function for CNF with different architectures and for Attack and Aggression.} 
    \label{fig:cnf_histo}
\end{figure}

For the combined set of Aggression \& Attack, visualizations of the waveform of the probability function were prepared as a result of Experiment 3. described in the publication, are presented in Fig.~\ref{fig:realnvp_histo} and Fig.~\ref{fig:cnf_histo}.

\end{document}